\documentclass[sn-apa]{sn-jnl}

\usepackage{graphicx}%
\usepackage{multirow}%
\usepackage{amsmath,amssymb,amsfonts}%
\usepackage{amsthm}%
\usepackage{mathrsfs}%
\usepackage[title]{appendix}%
\usepackage{xcolor}%
\usepackage{textcomp}%
\usepackage{manyfoot}%
\usepackage{booktabs}%
\usepackage{algorithm}%
\usepackage{algorithmicx}%
\usepackage{algpseudocode}%
\usepackage{listings}%
\usepackage{tabularx}
\usepackage{subcaption} 
\usepackage{graphicx}
\usepackage{numprint} 
\usepackage{float} 

\begin{document}

\title[Article Title]{Want a Ride? Attitudes Towards Autonomous Driving and Behavior in Autonomous Vehicles}


\author[]{\fnm{Enrico} \sur{Del Re}}\email{enrico.del\_re@jku.at}

\author[]{\fnm{Leonie} \sur{Sauer}}\email{k12220384@students.jku.at}

\author[]{\fnm{Marco} \sur{Polli}}\email{k01605988@students.jku.at}

\author*[]{\fnm{Cristina} \sur{Olaverri-Monreal}}\email{cristina.olaverri-monreal@jku.at}

\affil[]{\orgdiv{Department Intelligent Transport Systems}, \orgname{Johannes Kepler University Linz}, \orgaddress{\street{Altenbergerstr. 69}, \city{Linz}, \postcode{4040}, \state{Upper Austria}, \country{Austria}}}


\abstract{Research conducted previously has focused on either attitudes toward or behaviors associated with autonomous driving. In this paper, we bridge these two dimensions by exploring how attitudes towards autonomous driving influence behavior in an autonomous car. We conducted a field experiment with twelve participants engaged in non-driving related tasks. Our findings indicate that attitudes towards autonomous driving do not affect participants' driving interventions in vehicle control and eye glance behavior. Therefore, studies on autonomous driving technology lacking field tests might be unreliable for assessing the potential behaviors, attitudes, and acceptance of autonomous vehicles.}




\maketitle

\section{Introduction}
\label{sec:introduction}

With the introduction of autonomous vehicles on the streets, exemplified by driverless taxis from companies like Cruise and Waymo in San Francisco, the conversation around autonomous cars has evolved from futuristic speculation to a pressing issue in contemporary society~\cite{panasewicz2023perspective}. Public discourse, shaped by personal experiences with existing automation technologies (e.g., adaptive cruise control, lane-keeping assistance), media coverage, and the automotive industry, has emerged to discuss both the opportunities and challenges of widespread adoption of autonomous vehicles. While some anticipate increased efficiency, reduced risk, and enhanced accessibility and convenience, others express concerns about issues such as data security, loss of human control, and skill deterioration~\cite{fraedrich2016societal, olaverri2016autonomous, peng2020ideological, saffarian2012automated}. Consequently, various stakeholders, including policymakers, are keenly interested in understanding attitudes towards autonomous driving and human-machine interaction in autonomous vehicles. This interest aims to align current mobility strategies with people's needs and expectations~\cite{rahimi2020examining}, \cite{ward2017acceptance}.

From a sociological standpoint, mobility and mobility behavior are complex subjects intertwined with various dimensions, such as socio-demographics (e.g., rural or urban residence), technology acceptance (e.g., preference for driving cars), and lifestyle choices (e.g., environmentally conscious living)~\cite{herrenkind2019young}. As we transition from manual to autonomous driving, these factors and their correlations may shift due to changes in human-machine interaction. This includes differences in tasks, roles, and control for both the autonomous vehicle and the human occupant, as well as evolving levels of trust in the technology and new perceptions and expectations regarding driving and traffic~\cite{certad2023extraction, olaverri2020promoting, nyholm2020automated}. Therefore, a comprehensive understanding of autonomous driving interaction requires consideration of these multiple factors.

Several studies have explored autonomous driving within the context of sociotechnical transformation processes, aiming to understand individuals' perceptions of self-driving cars and their willingness to adopt them. For instance, ~\cite{fraedrich2016societal} highlighted that acceptance is shaped by factors at both individual and societal levels. They further differentiated between:\\
Acceptance subject that encompasses the driver or passenger and includes attitudes, judgments, motivations, car usage, ownership, comfort, trust or skepticism, freedom or control, and enjoyment of driving.\\
Acceptance object that refers to the car itself, encompassing perceived technological features, development potential, legal frameworks, liability issues, safety considerations, implications for transportation and social systems, flexibility, comfort, efficiency, and uncertainty.

In our study, which focused specifically on attitudes toward autonomous vehicles (pertaining to the acceptance subject), we identified relevant variables. Additionally, we explored the correlation between attitudes toward autonomous driving and the behavior of individuals within an autonomous vehicle. By integrating these two dimensions into a single research design, we address a significant gap in existing literature. This user- and usage-oriented approach aimed to produce findings that are relevant and actionable for policymakers and vehicle developers.

To this end, we defined the following null hypotheses:

$H0_1$: There is no correlation between attitudes towards autonomous driving and the frequency of driving interventions made by individuals while riding in autonomous vehicles.

$H0_2$: There is no correlation between attitudes towards autonomous driving and individuals' eye glance behavior while riding in an autonomous vehicle.

The next section presents theoretical insights on attitudes and human-machine interaction in the field. Sections~\ref{sec:method} and~\ref{sec:data} outline the methodology employed to collect and analyze the relevant data. Section~\ref{sec:results} presents the findings, while Section~\ref{sec:conclusion} discusses and concludes the paper proposing future research.

\section{Related literature}
\label{sec:related}
Human-Robot Interaction (HRI) and Human-Machine Interaction (HMI) encompass the relational and mutual actions and communications between at least one human and one machine. They also involve the division of tasks, roles, and control~\cite{both2013hands}. The dynamic nature of HMIs in automotive contexts has been emphasized in several studies~\cite{olaverri2016human}, reshaping the driver's role to that of a supervisor. However, when a vehicle reaches its Operational Design Domain (ODD) limit, a driver response is expected. In such cases, cooperation between the automated driving system and the driver can ensure a safe and comfortable transition, facilitated, for example, by a haptic guidance system~\cite{morales2022automated}.

Cooperative actions also occur in traffic as human drivers initiate socially negotiated behaviors such as eye contact, gestures and other to signal future actions.

The introduction of autonomous vehicles raises research questions about interactions between drivers of vehicles with varying degrees of automation, as highly automated vehicles may not be able to interpret non-verbal cues from other road users. Developing algorithms for the automatic analysis of pedestrian body language can provide insights into how different interaction paradigms affect public perceptions of road safety and trust in new vehicular technologies~\cite{alvarez2019response}. 

Furthermore, Software as a Service (SaaS) platforms offer information exchange and cloud computing services in this context~\cite{smirnov2021game, smirnov2022interaction}.

The transformation towards vehicle automation presents challenges related to understanding human driving behavior and what drivers perceive as safe. Bridging the gap between subjective perceptions and objective safety measures has been the focus of numerous studies in specific scenarios~\cite{del2022implementation}.

Regarding the interaction between passengers and autonomous vehicles, processes of trust-building are required to make sure that humans can be engage in various activities such as reading newspapers, eating, or using their phones. This paper presents findings regarding human behavior in autonomous vehicles, examining how this behavior correlates with their attitudes toward autonomous vehicles.

\section{Experimental Setup}
\label{sec:method}

A driving experiment was carried out in accordance with the relevant guidelines and regulations. Informed consent was obtained from all participants.

To collect attitudes toward autonomous driving from the participants, immediately prior to the field experiment, they filled out a pre-task questionnaire on-site. It based on a standardized quantitative survey to differentiate between participants based on their varying experiences, rather than differences stemming from the context of the questioning~\cite{reinecke2019grundlagen}. 

The questionnaire started with a segment on attitudes towards autonomous driving, created based on related literature~\cite{pew2017}. It comprised nine items, exploring knowledge about autonomous vehicles, emotional responses (such as enthusiasm, concern, and perceived safety), and attitudes towards them, including willingness to drive one, regulatory views, and potential consequences. Each item presented statements and utilized a four-point scale for agreement, with an additional option for no answer.
This was followed by a general demographic section.\\

The sample consisted of 12 participants with valid driver's licenses, ensuring a diverse representation across socio-demographic factors. The group comprised 5 male and 7 female participants, with age categorized into five groups. The majority of respondents were in the younger three age groups. \\



\begin{figure}[!t]
	\centering
	\includegraphics[width=0.5\textwidth]{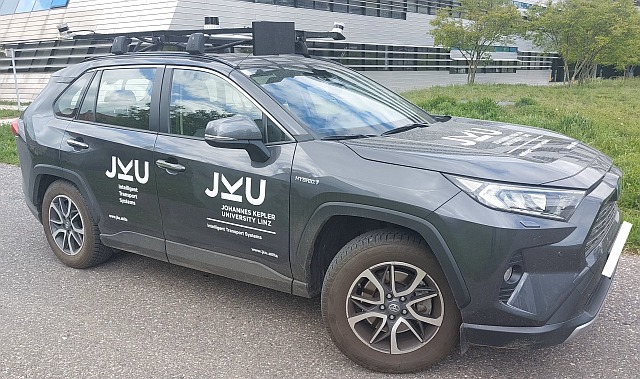}
	\caption{Research vehicle from the Intelligent Transport Systems Department at the Johannes Kepler University Linz in Austria. The vehicle was utilized to perform the field test experiments. }
	\label{fig:CarJKU-ITSk}
\end{figure}

The experimental setup was based on the design from~\cite{morales2023transferability}. Data collection was conducted using the JKU-ITS vehicle (see Figure~\ref{fig:CarJKU-ITSk}), as detailed in \cite{certad2023jku}] and ~\citep{barrio2023development}. During the experiment, participants performed various tasks while the vehicle autonomously navigated a predetermined test lane on the university campus.\\

\begin{figure}[!t]
	\centering
	\includegraphics[width=0.5\textwidth]{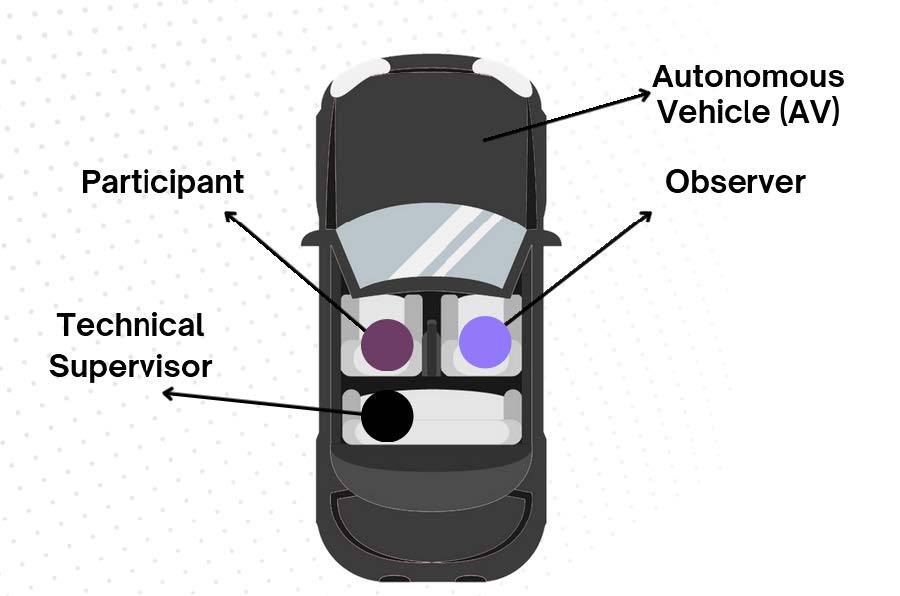}
	\caption{Seating arrangement in the JKU-ITS vehicle during testing }
	\label{fig:experimentalsetupcar}
\end{figure}

Participants sat behind the steering wheel, with a member of the research team in the passenger seat. A technical supervisor occupied the back seat, overseeing an override system capable of intervening in the autonomous driving if necessary. Figure~\ref{fig:experimentalsetupcar} illustrates the seating arrangement in the vehicle.

Prior to the experiment, participants received an introduction to the car's technological specifications and safety features from the technical supervisor.

To start the experiment, participants manually drove the car to a predetermined starting point. At this location, the autonomous driving system was activated.

As the car autonomously navigated the designated route, participants were tasked with performing seven different activities, outlined by the technical supervisor. These activities are listed in Table~\ref{tab:activities}.

\begin{table}[ht]
    \caption{List of non-driving related tasks performed during the field test}
    \label{tab:activities}
    \centering
    \begin{tabular}{|c|l|}
        \hline
        \textbf{Activity Number} & \textbf{Activity} \\
        \hline
        1 & Reading a newspaper \\
        \hline
        2 & Reading a magazine \\
        \hline
        3 & Reading a book \\
        \hline
        4 & Drinking from a bottle \\
        \hline
        5 & Talking on the phone \\
        \hline
        6 & Manipulating a smartphone \\
        \hline
        7 & Observing the surroundings \\
        \hline
    \end{tabular}
\end{table}

The order of activities was randomized. After each task on the testing route, participants deactivated the autonomous mode, manually turned the car around, and then re-engaged the autonomous mode to proceed with the next task in the opposite direction. 

\section{Data acquisition and analysis}
\label{sec:data}
Experimental observations were annotated, with particular attention to instances where the driver intervened in the control of the car by touching the steering wheel or using the pedals. We also recorded the participants' gaze direction (front window, mirrors, etc.) and the level of task completion (fully completed (1 point), partially completed (0.5 points), or not completed (0 points)). The objective was to assess whether participants could focus on other tasks while riding in an autonomous car. For instance, taking an extended pause from reading to observe the road was categorized as a partially completed reading task. \\

Additionally, we documented the sequence of tasks performed. When a driver intervention occurred, we further registered whether it was deemed necessary or not,  The necessity was determined based on whether the car was driving smoothly without any external disruptions. For example, one participant pressed the brakes when a pedestrian approached a crossing. This intervention was categorized as unnecessary since the car would have naturally slowed down. On the other hand, another participant corrected the steering wheel when the car deviated from its intended path, approaching an obstacle on the left. This intervention was classified as necessary. 
Lastly, we annotated and categorized participant comments.\\
 
After concluding the experiment, the observational notes were matched with the survey responses.
We examined correlations between the task score, frequency of participant intervention (intervention score), frequency of looking up from the windshield (glance score), and frequency of looking in the mirrors (mirror score) with the scores from the pre-task questionnaire. \\

To quantify the documented observations, we assigned a task completion score to each participant. The maximum score achievable, assuming all tasks were fully completed, was 7 points.

For other observed behaviors, we used binary variables to indicate their occurrence: 1 for a particular observed behavior and 0 when it did not occur.

Regarding the pre-task questionnaire, we quantified the findings by assigning scores to the attitudinal responses and then calculated the average scores based on Likert-scale from 1 to 4. Statements left unanswered were excluded from the analysis. 
We established threes distinct scores to represent the attitudes explored in the questionnaire:

\begin{itemize}
\item Emotionality Score: This score indicated whether participants had a generally positive (1 point) or negative (4 points) emotional response to autonomous driving. The items for this score
were:\\
``How enthusiastic are you about the development of driverless vehicles?''\\
``How worried are you about the development of driverless vehicles?''

\item  Safety Score: This score reflected the participants' subjective safety when interacting with driverless vehicles, with 1 point for feeling completely safe and 4 points for feeling completely unsafe, assessed with the following items:\\
``How safe would you feel sharing the road with a driverless passenger vehicle?''\\
``How safe would you feel sharing the road with a driverless freight truck?''

\item Rules Score: This score determined the participants' perception regarding the necessity of a legal framework for autonomous vehicles, with 1 point standing for rejecting regulations and 4 points for welcoming them. Below are the suggested regulations:\\
``Regulation requiring driverless vehicles to operate in special lanes.''\\
``Regulation prohibiting driverless vehicles from operating near certain areas, such as schools.'' \\
``Requirement for a person capable of taking over in an emergency to be seated in the driver’s seat''.

\end{itemize}

To analyze the influence of the independent variables (attitudes) on the dependent variables (task completion and behavior), the Pearson correlation coefficient was calculated for task completion and the Point-Biserial correlation coefficient for the behaviors with a significance of $p\le 0.05$. Furthermore, a regression analysis was conducted  to obtain further insights into the connection between dependent and independent variables.
    
\section{Results}
\label{sec:results}

\begin{figure}[!t]
	\centering
	\includegraphics[width=0.6\textwidth]{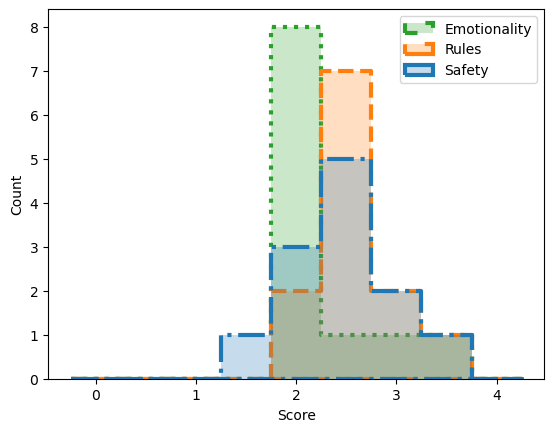}
	\caption{Distribution of the answers to the questionnaire for each category: Emotional Score, Regulation Score and Safety score.}
	\label{fig:independent_hist}
\end{figure}

The distributions of attitude scores for the 12 participants are shown in Figure \ref{fig:independent_hist} as histograms. The values are grouped close to the neutral answer score of 2.5 and shown in more detail in Table \ref{tab:quest_res}. A Student's t-test with the null hypothesis of a neutral attitude $\mu_0 = 2.5$, a two-sided alternative, and a confidence level of $p-value<0.05$ resulted in no rejection of the null hypothesis.

The largest difference to a neutral attitude can be observed for the Safety Score, though it is not statistically significant.

\begin{table}[ht]
    \centering
        \caption{The results from the scores for the attitudes evaluated through the questionnaires of 12 participants and the corresponding t-test. No distribution differed significantly from a distribution centered around a neutral attitude score of $2.5$}
    \label{tab:quest_res}
    \npdecimalsign{.}
    \nprounddigits{2}
    \begin{tabular}{l|n{5}{2}|n{5}{2}|n{5}{2}|n{5}{2}}
    \toprule
    \multicolumn{1}{c|}{$N=12$}& \multicolumn{1}{c|}{mean} & \multicolumn{1}{c|}{std} & \multicolumn{1}{c|}{t-test} & \multicolumn{1}{c}{p-value}\\ \hline
    Emotionality Score &   $2.458333$   &  $0.541812$    &    $-0.2663977138330646$    &     $0.7948604511751455$    \\ \hline 
    Rules Score        &   $2.555556$   &   $0.404104$   &   $0.4762395649758403$     &      $0.6432226072577636$   \\ \hline 
    Safety Score       &   $2.272727$   &   $0.517863$   &   $-1.4555562743489538$     &    $0.17617927161701802$  \\ \bottomrule
    \end{tabular}
    \npnoround
\end{table}

\begin{figure}[htbp]
    \centering
    \begin{subfigure}[b]{0.45\textwidth}
        \includegraphics[width=\textwidth]{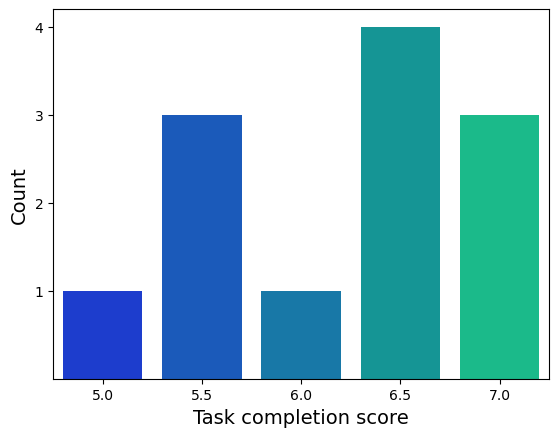}
        \caption{Counts of Task Completion Scores.}
        \label{fig:taskscount}
    \end{subfigure}
    \hfill
    \begin{subfigure}[b]{0.45\textwidth}
        \includegraphics[width=\textwidth]{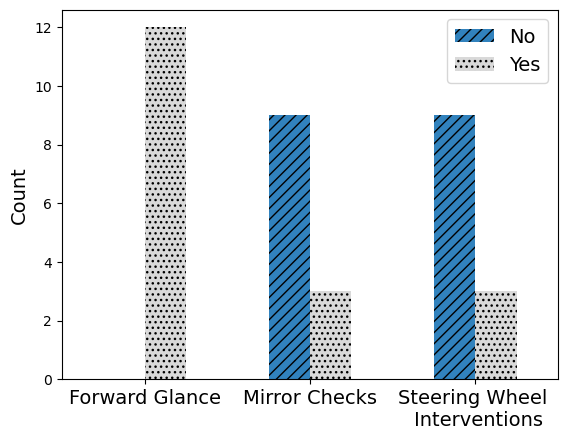}
        \caption{Count of observed behaviors.}
        \label{fig:behaviors}
    \end{subfigure}
    \caption{Results of the dependent variables, Task Completion Scores and observed behaviors. The mean Task Completion Score is $6.21$ and the standard deviation is $0.69$.}
    \label{fig:Task_Behavior}
\end{figure}

The distribution of the Task Completion Scores (see Figure \ref{fig:taskscount}) revealed that some participants achieved a full score, being the lowest score 5 points. However, no task was fully completed by all participants.

The observed behavior during the task completion is shown in Figure \ref{fig:behaviors}. Forward Glances were observed for all participants, only $25\%$ of them checked the mirrors or intervened with the steering wheel.

\begin{figure}[!t]
	\centering
	\includegraphics[width=0.8\textwidth]{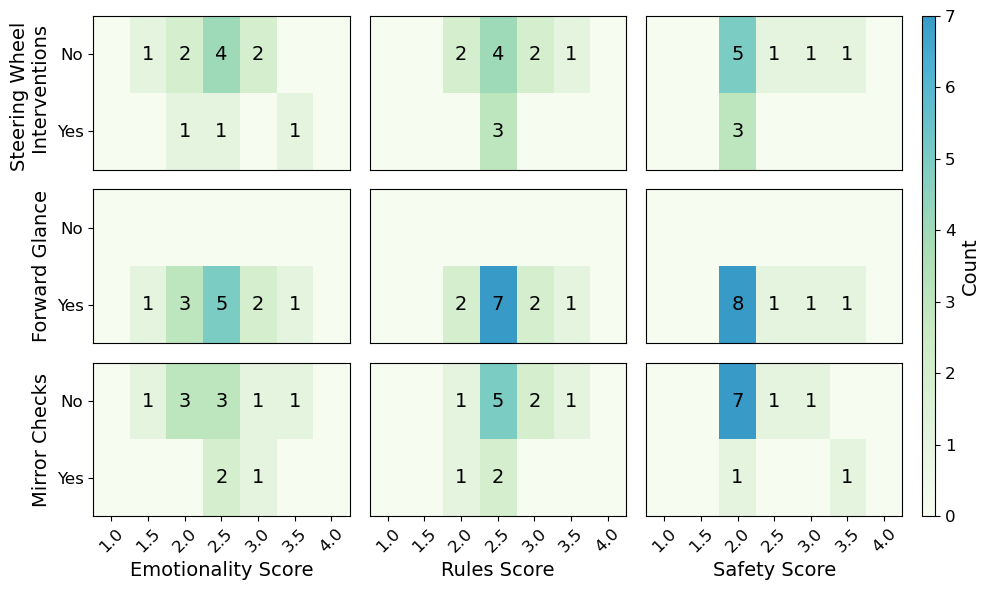}
	\caption{2D-histogram of the bivariate distributions of the attitude scores and observed behaviors with counts of occurences.}
	\label{fig:correlations}
\end{figure}

The bivariate distributions of the attitude scores and the observed behaviors are shown in Figure \ref{fig:correlations}. No correlation is directly visible, which is confirmed through pairwise point-biserial correlation tests yielding only non-significant results. Additionally, multiple logistic regression using the three attitudes as input variables results in only non-significant coefficients for the attitude scores.

\begin{figure}[!t]
	\centering
	\includegraphics[width=0.8\textwidth]{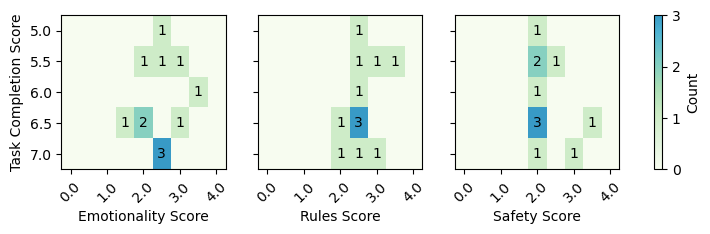}
	\caption{2D-histogram of the bivariate distributions of the attitude scores and Task Completion Score with counts of occurences.}
	\label{fig:tasks}
\end{figure}

Similarly, the bivariate distribution of attitude scores and the Task Completion Score shown in Figure \ref{fig:tasks} does not indicate any correlation. Pairwise Pearson correlation tests confirm this observation as no statistically significant correlations are found. Likewise, multiple linear regression using ordinary least squares results in non-significant coefficients for the three attitudes.

Thus, due to the lack of statistically significant correlation coefficients, the defined hypotheses H01 and H02 are both accepted. Especially for H02 the correlation was not calculable due to uniform behavior across all participants. 
Regression analysis further emphasized the lack of correlation, though a larger sample size would be necessary to perform further tests and strengthen the statements of the ones used above.

The comments from the participants during the experiment related to emotional aspects such as trust, control, and feelings of safety, as well as practical considerations like the mobility experience, in-car behavior, and alternative transportation modes. They were generally positive, with interventions and mirror checks being only occasional. Some of the comments were: 
``Reading would be practical but difficult on complicated road situations.'' 
        ``It is kind of strange to give the car control over driving.'' 
                ``I behave regularly out of habit.'' 
        ``Feels like sitting in the bus (...) or taxi.'' 
        ``I'm more afraid of automatic functions than autonomous driving.''    
        ``Driving with automatic functionality causes more problems for me than autonomous driving.'' 
        ``Very focused'' (referred to own task performance).
        ``Yes, there was a child!'' (being asked whether the participant was scared). 
        ``Now it overlooked the pedestrian? really!'' (during phone call)".

 \section{Conclusion and future work}
 \label{sec:conclusion}
The existing literature presents diverse approaches to examining autonomous driving from a sociological perspective. However, the majority of studies focus solely on either attitude toward autonomous vehicles or behaviour within these vehicles. 
This study focused on integrating the two dimensions into a single research design.

Results indicated that there is great potential in employing a two-folded research design to explore autonomous driving. No statistically significant correlation between attitudes towards autonomous driving with both individuals' driving interventions and eye glance behavior while riding autonomous cars were identified. This finding suggests that questionnaires and surveys on attitudes and potential behavior towards autonomous driving technology may not necessarily correlate with real-world behavior in road situations. Therefore, to draw accurate conclusions, it is crucial to always conduct field tests. 

The experiment was conducted in a controlled environment on a test track without other vehicles, prioritizing safety for the test drivers. As a result, certain observations should be interpreted with caution. For instance, the lack of checks in the side and rearview mirrors could be due to the absence of other cars on the road.

While our study offers valuable insights, there are limitations that may affect the generalizability of the results, such as the small sample size and the specific demographic characteristics of the participants. Therefore, future research will consider a larger sample.


=============================================%

\bibliography{references}

\end{document}